\documentclass[twocolumn,10pt]{article}

\usepackage[utf8]{inputenc}
\usepackage[T1]{fontenc}
\usepackage{lmodern}
\usepackage[margin=1.9cm]{geometry}
\usepackage{amsmath,amssymb,amsfonts}
\usepackage{graphicx}
\usepackage{booktabs}
\usepackage{array}
\usepackage{xcolor}
\usepackage{tikz}
\usetikzlibrary{arrows.meta,positioning,calc,shapes.geometric}
\usepackage{caption}
\usepackage{enumitem}
\usepackage[hidelinks]{hyperref}
\usepackage{dblfloatfix} 

\graphicspath{{figures/}}

\newcommand{\ind}{\mathbf{1}}
\newcommand{\lse}{\mathop{\mathrm{logsumexp}}}

\title{\bfseries Differentiable Clone-Structured Causal Graphs for\\ End-to-End Cognitive Map Learning from Image Sequences}

\begin{document}

\twocolumn[
\begin{@twocolumnfalse}
\begin{center}
{\LARGE\bfseries Differentiable Clone-Structured Causal Graphs for\\ End-to-End Cognitive Map Learning from Image Sequences\par}
\vspace{0.9em}
{\large Arash Nikzad\textsuperscript{1}, Sasan Sarbishegi\textsuperscript{2}, Ali Dasmeh\textsuperscript{3}, Muhammad Asif\textsuperscript{4}, Parsa Gharavi\textsuperscript{1}, Erik Husom\textsuperscript{5}, Sagar Sen\textsuperscript{5}, Andrew B. Lehr\textsuperscript{6,7,11}, Olivier Penacchio\textsuperscript{8,9}, Ana Clemente\textsuperscript{4}, Tristan M. St\"ober\textsuperscript{7,10,11,*}\par}
\vspace{0.6em}
{\footnotesize
\textsuperscript{1}Goethe University Frankfurt, Frankfurt, Germany;\;
\textsuperscript{2}Independent Researcher, Teheran, Iran;\;
\textsuperscript{3}Max Planck Institute for Human Development, Berlin, Germany;\;
\textsuperscript{4}Department of Cognitive Neuropsychology, Max Planck Institute for Empirical Aesthetics, Frankfurt, Germany;\;
\textsuperscript{5}SINTEF, Oslo, Norway;\;
\textsuperscript{6}Department of Neuro- and Sensory Physiology, University Medical Center G\"ottingen, G\"ottingen, Germany;\;
\textsuperscript{7}Institute of Computer Science and Campus Institute Data Science, University G\"ottingen, Germany;\;
\textsuperscript{8}Bridging Research in AI and Neuroscience (brAIN), Computer Vision Center, Bellaterra, Spain;\;
\textsuperscript{9}Computer Science Department, Universitat Aut\`onoma de Barcelona, Bellaterra, Spain;\;
\textsuperscript{10}Epilepsy Center Frankfurt Rhine-Main, Department of Neurology, Goethe University Frankfurt, Frankfurt, Germany;\;
\textsuperscript{11}Circulant Labs, Bensheim, Germany.\\[0.4em]
\textsuperscript{*}Corresponding author : tristan.stoeber@posteo.net\par}
\end{center}
\vspace{0.8em}
\begin{center}
\begin{minipage}{0.86\textwidth}
\small
\textbf{Abstract.}
How can an agent build a structured map of its world from nothing but an ongoing sequence of raw sensory input and its own movements, especially when natural variation means exact sensory patterns rarely repeat?
The Clone-Structured Causal Graph algorithm (CSCG), a normative hippocampus model, shows how an interpretable map can be learned from aliased observations.
However, CSCG requires a predefined discrete alphabet, and its expectation-maximization formulation is not easily combined with existing neural network modules, preventing the end-to-end processing of raw image sequences.
We remove this barrier by reformulating CSCG as a single, fully differentiable module, gradCSCG, and coupling it to a learned vector-quantized variational autoencoder (VQ-VAE) perceptual front-end.
A soft emission forward pass allows the map-learning objective to flow back into perception, while a set of loss-balancing mechanisms mitigates module collapse during joint training.
We demonstrate, first, that gradient training reproduces CSCG's results on original symbolic grid worlds by recovering room topology from heavily aliased observations.
Second, we show that map recovery remains robust on MNIST image sequences, where each visit to a location yields a newly sampled image of its assigned digit.
Across four heavily aliased environments, the end-to-end pipeline successfully uncovers the underlying adjacency graph with high edge precision and recall, directly from visual input.
This work provides a proof of principle that CSCG can serve as a composable building block in a deep learning architecture.
\par\vspace{6pt}\noindent
\end{minipage}
\end{center}
\vspace{1.2em}
\end{@twocolumnfalse}
]

\section{Introduction}
Well-structured internal representations allow biological and artificial agents to pick shortcuts on routes never taken and to provide guidance in situations where trial-and-error learning would be fatal.
Thus, understanding and reengineering the emergence of such representations is a fundamental research frontier both in neuroscience [1--5] and artificial intelligence (AI) [6--8].

The \textbf{Clone-Structured Causal Graph} (CSCG) algorithm [9, 10], a normative hippocampus model, explains how well-structured representations can emerge from experience.
Technically, CSCG is an overcomplete hidden Markov model (HMM) with a fixed emission matrix that creates a statistical model from sequential observations.
Compressing a series of observation--action pairs into a higher-order representation of its environment, CSCG learns to disambiguate contexts from aliased observations.
Obeying the Markov property --- i.e.\ any subsequent state depends only on the current state --- CSCG is forced to represent a novel context with a new clone among its hidden nodes.
The graph emerging from this cloning operation provides a condensed map of the environment and is suitable for planning, consolidation and abstraction.
However, while elegantly creating well-structured representations in static and relatively small environments, it is unclear how to extend this approach to richer, perceptually complex observations.
The bottleneck is that CSCG is trained by Expectation--Maximization over a fixed, discrete observation alphabet, which prevents seamless composition with gradient-trained neural modules.

We resolve this by reimplementing CSCG as a single differentiable, gradient-trained computation [11] in TensorFlow --- a model we call \textbf{gradCSCG}. This innovation allows us to co-train gradCSCG with a vector-quantized variational autoencoder (VQ-VAE).
We demonstrate that this approach preserves CSCG's expressivity while enabling it to handle sensory variability and complexity in environments composed of MNIST digits.

\noindent\textbf{Contributions}
\begin{enumerate}[leftmargin=1.4em,itemsep=2pt,topsep=2pt]
\item An \textbf{end-to-end trainable pipeline composed of a gradient-based CSGG and a VQ-VAE} able to create a topological map from sequences of images.
\item \textbf{Loss-balancing for stable joint training} --- length normalization, weight annealing, a diversity penalty, and anti-collapse safeguards (Section 3.6).
\item A formal, reusable \textbf{topology-recovery evaluation suite} (Section 3.10).
\item An empirical study on four MNIST grid-world environments with strong aliasing (Sections 4--5).
\end{enumerate}

\subsection{Related Work}
\textbf{CSCG toolkit.} A differentiable cloned-HMM forward pass with a soft-observation interface and encoder-gradient flow has also been developed in concurrent, independent open-source work [21], whose image experiment couples a convolutional network with a gradient-based CSCG for supervised digit classification. Relative to this work, our specific contributions are the learned VQ-VAE discretizer trained jointly with the sequence model, the loss-balancing that keeps that joint training stable, and --- crucially --- the use of this gradient-based setup to \textit{actually recover and evaluate an environment's map}: we deduce the physical adjacency graph from the learned transitions and score it against the ground-truth topology (Section 5). Their image experiment, by contrast, uses a convolutional--CSCG coupling for supervised digit classification only. They do not attempt to recover the topology, which is the central question our pipeline is built to answer.

\textbf{Neural sequence models for cognitive maps.} A complementary line learns maps from the latent codes of a neural predictor rather than from a cloned HMM. Dedieu et al.\ [22] train a transformer with discrete bottlenecks on next-observation prediction and read interpretable cognitive maps off its bottleneck indices for planning in partially observed environments. As in our pipeline, a neural representation is discretized and a map is recovered from observation--action streams; the difference is \textit{where the map lives}. In their approach the cognitive map is decoded from the transformer's bottleneck codes as a separate, post-hoc analysis and handed to an external solver for planning, so the network's own working representation stays dense and is not itself an interpretable, directly usable map. In ours the clone-graph sequence model is co-trained with the perceptual discretizer and its transition matrix \textit{is} the map: an intrinsically interpretable, on-line structure that the model computes with and can be queried directly for planning and for understanding the environment.

\textbf{Dynamic and expanding maps.} Closer to the open problems we raise in Section 7, de Tinguy et al.\ [23] grow a cognitive map online --- dynamically expanding it over predicted poses within an active-inference agent --- and benchmark against CSCG on grid environments. Their focus on a map that \textit{expands} as the agent explores is precisely the dynamic-allocation capability that our fixed clone budget lacks.
However, their model learns via discrete counting operations that are not natively differentiable, making it difficult to couple seamlessly with deep neural networks for visual processing. By contrast, our fully differentiable formulation allows the map to be co-trained with a neural front-end, learning the discretization end-to-end directly from raw pixels.

\section{Methods}

\subsection{Problem formulation}
An agent produces an episode of length $T$: observations $x_{1:T}$ with $x_t\in\mathcal{X}\subset\mathbb{R}^{H\times W\times C}$ and actions $a_{1:T-1}$ with $a_t\in\mathcal{A}=\{1,\dots,A\}$, where $a_t$ is taken between times $t$ and $t{+}1$. Each observation is emitted at an underlying physical place $g_t\in\mathcal{G}$; the environment has a true undirected adjacency graph $\mathcal{M}=(\mathcal{G},\mathcal{E})$. The places $g_t$ and edges $\mathcal{E}$ are used \textit{only for evaluation} and are never seen during training. The goal is to learn, from $(x_{1:T},a_{1:T-1})$ alone, a latent model whose transition structure recovers $\mathcal{M}$.

The pipeline (Figure 1) has two modules: a VQ-VAE that maps each image to a discrete token, and an action-conditioned cloned HMM over those tokens whose latent graph is the learned map.

\begin{figure*}[t]
\centering
\begin{tikzpicture}[
  >={Latex[length=2.4mm]},
  box/.style={draw, rounded corners=2pt, align=center, minimum height=8mm,
    inner xsep=3pt, inner ysep=3pt, font=\small},
  loss/.style={draw, ellipse, align=center, inner sep=2pt, font=\small, fill=black!5},
  fwd/.style={->, thick},
  grad/.style={->, dashed, gray}]
  \node[box] (img) {Image $x_t$};
  \node[box, right=10mm of img] (enc) {Encoder $E_\phi$};
  \node[box, above right=6mm and 12mm of enc] (vq) {Codebook $\{e_k\}$};
  \node[box, right=12mm of vq] (dec) {Decoder $D_\psi$};
  \node[loss, right=12mm of dec] (rec) {$\mathcal{L}_{\mathrm{rec}}$};
  \node[box, below right=6mm and 12mm of enc] (soft) {Soft posterior $\rho_t$\\ over $K$ codes};
  \node[box, right=12mm of soft] (hmm) {gradCSCG\\ forward (soft)};
  \node[loss, right=12mm of hmm] (nll) {$\mathcal{L}_{\mathrm{gradCSCG}}$};
  \draw[fwd] (img) -- (enc);
  \draw[fwd] (enc) -- node[above,font=\scriptsize,sloped] {$z_t$} (vq);
  \draw[fwd] (vq) -- node[above,font=\scriptsize] {$\tilde q_t$} (dec);
  \draw[fwd] (dec) -- (rec);
  \draw[fwd] (enc) -- (soft);
  \draw[fwd] (soft) -- node[above,font=\scriptsize] {$\log\rho_t$} (hmm);
  \draw[fwd] (hmm) -- (nll);
  \draw[grad] (rec) to[bend right=12] (dec);
  \draw[grad] (nll) to[bend left=12] (hmm);
  \draw[grad] (hmm) to[bend left=12] (soft);
  \draw[grad] (soft) to[bend left=10] (enc);
\end{tikzpicture}
\caption{\textbf{The gradCSCG pipeline.} Solid arrows: forward computation. Dashed arrows: gradient flow. The encoder feeds both a reconstruction branch (hard quantization $\tilde q_t$ + decoder) and a sequence branch (soft codebook posterior $\rho_t$ + differentiable cloned-HMM forward pass). Because the gradCSCG likelihood is differentiable in $\rho_t$, the topological objective $\mathcal{L}_{\mathrm{gradCSCG}}$ shapes the encoder. The codebook itself is updated by an Exponential Moving Average (EMA), not by gradients.}
\end{figure*}
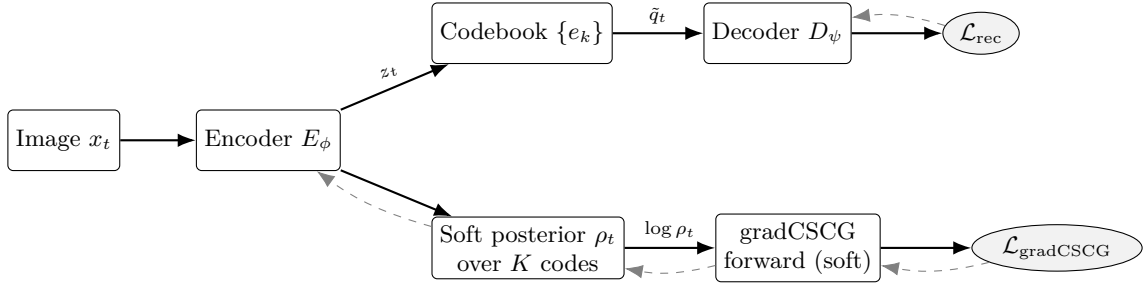

\subsection{Perceptual front-end: VQ-VAE}
\textbf{Encoder and quantization.}
A convolutional encoder $E_\phi:\mathcal{X}\to\mathbb{R}^{D}$ maps each image to a latent $z_t=E_\phi(x_t)$. A codebook $\{e_k\}_{k=1}^{K}$, $e_k\in\mathbb{R}^{D}$, where $K$ denotes the number of discrete latent codes and $D$ is the dimensionality of each codebook embedding, defines a discrete token by nearest-neighbour assignment,
\[
k_t=\arg\min_{k\in\{1,\dots,K\}}\;\lVert z_t-e_k\rVert_2^2, \qquad q_t=e_{k_t},
\]
where the squared distance is computed as $\lVert z-e_k\rVert_2^2=\lVert z\rVert_2^2-2\langle z,e_k\rangle+\lVert e_k\rVert_2^2$. Gradients cross the non-differentiable $\arg\min$ by the straight-through estimator [18],
\[
\tilde q_t = z_t + \mathrm{sg}(q_t-z_t),
\]
where $\mathrm{sg}(\cdot)$ is the stop-gradient operator, so the forward value is $q_t$ while $\partial\tilde q_t/\partial z_t=I$. A decoder $D_\psi$ reconstructs $\hat x_t=D_\psi(\tilde q_t)$.

\textbf{Losses.}
Over a minibatch $\mathcal{B}$,
\[
\mathcal{L}_{\mathrm{rec}} = \tfrac{1}{|\mathcal{B}|}\sum_{t} \lVert x_t-\hat x_t\rVert_2^2,
\]
\[
\mathcal{L}_{\mathrm{commit}} = \tfrac{1}{|\mathcal{B}|}\sum_{t} \lVert \mathrm{sg}(q_t)-z_t\rVert_2^2,
\]
the second term pulling encoder outputs toward their assigned code.

\textbf{EMA codebook updates.}
The codebook is \textit{not} trained by gradient descent. With decay $\gamma\in(0,1)$, per batch we accumulate, for every code $k$, a cluster size $n_k$ and a vector sum $m_k$,
\[
\begin{aligned}
n_k &\leftarrow \gamma\,n_k+(1-\gamma)\sum_{t}\ind[k_t=k],\\
m_k &\leftarrow \gamma\,m_k+(1-\gamma)\sum_{t}\ind[k_t=k]\,z_t,
\end{aligned}
\]
and set $e_k\leftarrow m_k/\hat n_k$ with Laplace-smoothed size
\[
\hat n_k=\frac{n_k+\epsilon}{\left(\sum_{k'}n_{k'}\right)+K\epsilon}\left(\sum_{k'}n_{k'}\right).
\]

\textbf{Soft codebook posterior.}
For the differentiable coupling (Section 3.5) the encoder also emits a temperature-controlled posterior over the codebook,
\[
\log\rho_t(k) = \log\mathrm{softmax}_{k}\left(-\lVert z_t-e_k\rVert_2^2/\tau\right),
\]
which is differentiable in $z_t$. As $\tau\to0$, $\rho_t$ concentrates on $k_t$ and this recovers the hard assignment.

\subsection{Sequence model: action-conditioned cloned HMM}
\textbf{State space and clone structure.}
Each token $k$ is assigned $C_k\ge1$ \textit{clones} --- latent states that all emit token $k$ but participate in different transition contexts. With a trailing \textit{sink} state $\bot$, the state space is $\mathcal{S}=\{1,\dots,N\}$ with $N=1+\sum_{k=1}^{K}C_k$. The sink state is a special terminal state that does not correspond to any visual token; it is used to absorb probability mass at the end of a sequence and to make sequence termination explicit in the HMM formulation. A fixed map $\omega:\mathcal{S}\setminus\{\bot\}\to\{1,\dots,K\}$ gives the token each state emits; in the uniform case $C_k\equiv C$ and $\omega(s)=\lceil s/C\rceil$. Emissions are \textit{deterministic}:
\[
B_{s,o}=\ind[\omega(s)=o],\qquad \log B_{s,o}=\begin{cases}0,&\omega(s)=o \\ -\infty,&\text{otherwise,}\end{cases}
\]
and the sink emits no real token. Clones are exactly the mechanism that disambiguates aliasing: one token observed at two places is explained by two clones with distinct transition rows.

\textbf{Parameters.}
The model has initial-state logits $\pi\in\mathbb{R}^{N}$ and action-conditioned transition logits $\Theta\in\mathbb{R}^{A\times N\times N}$, yielding
\[
\bar\pi=\mathrm{softmax}(\pi),\qquad T_{a,i,j}=\mathrm{softmax}_{j}\left(\Theta_{a,i,\cdot}\right).
\]
All learning resides in $(\pi,\Theta)$; emissions are fixed.

\textbf{Forward likelihood.}
Writing $\lse_i u_i=\log\sum_i e^{u_i}$, the log-forward messages for an episode $(o_{1:T},a_{1:T-1})$ obey
\[
\begin{aligned}
\log\alpha_1(j) ={}& \log\bar\pi_j+\log B_{j,o_1},\\
\log\alpha_{t+1}(j) ={}& \lse_{i}\left[\log\alpha_t(i)+\log T_{a_t,i,j}\right] + \log B_{j,o_{t+1}},
\end{aligned}
\]
and the episode log-likelihood is $\ell(o_{1:T}\mid a_{1:T-1})=\lse_{j}\log\alpha_T(j)$. The training loss is the mean negative log-likelihood (NLL)
\[
\mathcal{L}_{\mathrm{gradCSCG}}=-\tfrac{1}{|\mathcal{B}|}\sum_{(o,a)\in\mathcal{B}}\ell(o_{1:T}\mid a_{1:T-1}).
\]

\subsection{Gradient-based training of the cloned HMM}
Unlike the classical EM training of cloned HMMs, we evaluate the forward recursion and NLL loss as a single differentiable, log-space computational graph (a masked time recursion over padded minibatches) and optimize $(\pi,\Theta)$ directly by stochastic gradient descent with Adam [19]. Log-space arithmetic with $\lse$ keeps the recursion numerically stable over long episodes. This gradient formulation is what makes the sequence model composable with a neural front-end.

\subsection{Differentiable soft-emission coupling}
To let the topological objective shape perception, we replace the hard emission term $\log B_{j,o_{t}}$ in the forward pass by the soft log-posterior of the token that state $j$ emits:
\[
\begin{aligned}
\log\alpha^{s}_1(j) ={}& \log\bar\pi_j+\log\rho_1\left(\omega(j)\right),\\
\log\alpha^{s}_{t+1}(j) ={}& \lse_{i}\left[\log\alpha^{s}_t(i)+\log T_{a_t,i,j}\right] + \log\rho_{t+1}\left(\omega(j)\right),
\end{aligned}
\]
with the sink assigned $\log\rho_t(\bot)=-\infty$. The resulting log-likelihood $\ell^{s}$ is differentiable in $\rho_{1:T}$ and hence, through the soft posterior, in the encoder parameters $\phi$. Two properties hold by construction:
\begin{enumerate}[leftmargin=1.4em,itemsep=2pt,topsep=2pt]
\item \textbf{Consistency.} If $\rho_t$ is the one-hot distribution on the observed token $o_t$, then $\log\rho_t(\omega(j))=\log B_{j,o_t}$ and the soft recursion reduces exactly to the hard forward pass. As $\tau\to0$ the soft pipeline thus recovers the hard pipeline.
\item \textbf{End-to-end differentiability.} Gradients of $\mathcal{L}_{\mathrm{gradCSCG}}$ propagate $\ell^{s}\!\to\!\rho\!\to\!z\!\to\!\phi$, so the encoder is trained, in part, to produce tokens that make the action-conditioned sequence \textit{explainable}.
\end{enumerate}

\subsection{Joint objective and loss balancing}
We train the model in three phases. First, the VQ-VAE is trained independently to initialize the visual encoder and decoder and obtain stable discrete representations. Second, the complete model, comprising both the VQ-VAE and gradCSCG, is trained jointly in an end-to-end manner with stochastic gradient descent. Third, the VQ-VAE is frozen and gradCSCG is fine-tuned using hard emissions, allowing the temporal model to refine its transition structure while operating on discrete visual assignments.

\textbf{Combined objective.}
At joint step $t$,
\[
\mathcal{L}_{\mathrm{joint}} = \mathcal{L}_{\mathrm{rec}} + \beta\,\mathcal{L}_{\mathrm{commit}} + \lambda_t\,\widetilde{\mathcal{L}}_{\mathrm{gradCSCG}} + \alpha_{\mathrm{div}}\,\mathcal{L}_{\mathrm{div}}.
\]

\textbf{Length normalization.}
The raw NLL grows as $O(T)$, which on long episodes dwarfs the $O(1)$ reconstruction term and collapses the codebook. We therefore use the per-step NLL
\[
\widetilde{\mathcal{L}}_{\mathrm{gradCSCG}}=-\tfrac{1}{|\mathcal{B}|}\sum \tfrac{1}{T}\,\ell^{s}(o_{1:T}\mid a_{1:T-1}).
\]

\textbf{Weight annealing.}
The sequence-loss weight is ramped linearly so the codebook stabilizes under reconstruction before topological pressure turns on:
\[
\lambda_t=\lambda\cdot\min\left(1,\;t/T_{\mathrm{anneal}}\right).
\]

\textbf{Diversity penalty.}
Let $\bar\rho(k)=\frac{1}{|\mathcal{B}|}\sum_t\rho_t(k)$ be mean codebook usage and $H(\bar\rho)=-\sum_k\bar\rho(k)\log\bar\rho(k)$ its entropy. The penalty
\[
\mathcal{L}_{\mathrm{div}}=\log K-H(\bar\rho)\;\ge\;0
\]
vanishes only at uniform usage and counteracts codebook collapse.

\textbf{Anti-collapse safeguards and finalization.}
During Phase 2 we monitor codebook perplexity and keep the highest-perplexity checkpoint; optional $\lambda$-throttling, rollback, and dead-code revival provide further protection. A short \textit{finalization} phase then freezes the encoder and refines $(\pi,\Theta)$ on hard tokens with the unnormalized loss, optionally with a transition-entropy regularizer $-\eta\sum_{a,i}H(T_{a,i,\cdot})$ to sharpen transition rows. The following algorithm summarizes one joint step.

\begin{center}
\fbox{\begin{minipage}{0.92\columnwidth}
\small
\textbf{Algorithm 1 --- One joint training step}\\[2pt]
\textbf{Require:} image chunk $x_{1:T}$, actions $a_{1:T-1}$, weights $\beta,\lambda_t,\alpha_{\mathrm{div}}$, temperature $\tau$
\begin{enumerate}[leftmargin=1.4em,itemsep=1.5pt,topsep=2pt]
\item $z_t \gets E_\phi(x_t)$ \quad\textit{(encode)}
\item $q_t,\tilde q_t \gets \mathrm{quantize}(z_t)$; update codebook by EMA
\item $\hat x_t \gets D_\psi(\tilde q_t)$;\; $\mathcal{L}_{\mathrm{rec}},\mathcal{L}_{\mathrm{commit}} \gets$ reconstruction and commitment losses (Section 3.2)
\item $\log\rho_t \gets \log\mathrm{softmax}(-\lVert z_t-e_\cdot\rVert^2/\tau)$
\item $\ell^{s}\gets$ soft forward pass (Section 3.5)
\item $\widetilde{\mathcal{L}}_{\mathrm{gradCSCG}}\gets -\ell^{s}/T$;\; $\mathcal{L}_{\mathrm{div}}\gets \log K-H(\bar\rho)$
\item $\mathcal{L}_{\mathrm{joint}}\gets$ combined objective (Section 3.6)
\item update $(\phi,\psi,\pi,\Theta)$ with Adam on $\nabla\mathcal{L}_{\mathrm{joint}}$
\end{enumerate}
\end{minipage}}
\end{center}

\subsection{Decoding}
The maximum-a-posteriori state path is obtained by the Viterbi recursion [15]:
\[
\begin{aligned}
\delta_1(j) &= \log\bar\pi_j+\log B_{j,o_1},\\
\delta_{t+1}(j) &= \max_{i}\left[\delta_t(i)+\log T_{a_t,i,j}\right] + \log B_{j,o_{t+1}},
\end{aligned}
\]
with backpointers $\psi_{t+1}(j)=\arg\max_i\left[\delta_t(i)+\log T_{a_t,i,j}\right]$ and traceback $s^\star_T=\arg\max_j\delta_T(j)$, $s^\star_t=\psi_{t+1}(s^\star_{t+1})$, giving the most likely hidden state at each time $t$. The decoded path $s^\star_{1:T}$ is the basis of all evaluation.

\subsection{Token compaction and clone allocation}
In Phase 1, codes never emitted by the trained encoder are pruned and the alphabet is relabelled to the active set. Clone counts $C_k$ may be uniform or set per observation \textit{by hand} (a fixed, manually chosen per-observation clone budget); the deterministic emission structure of Section 3.3 is unchanged. This is a static hyperparameter chosen before training, \textbf{not} a learned, on-demand allocation of clones.

\subsection{Evaluation metrics}
All metrics are computed from the decoded path $s^\star_{1:T}$ of a held-out episode together with the ground-truth places $g_{1:T}$ and edges $\mathcal{E}$ (used only here, never in training). Let $\mathcal{V}$ be the set of visited states and $n(s,g)=\sharp\{t:s^\star_t=s,\;g_t=g\}$.

\textbf{State-to-place assignment.}
Each visited state is mapped to its majority place, $\chi(s)=\arg\max_{g}n(s,g)$.

\textbf{Clone purity.}
\[
\mathrm{Purity}=\frac{1}{|\mathcal{V}|}\sum_{s\in\mathcal{V}}\frac{\max_g n(s,g)}{\sum_g n(s,g)},
\]
with the visit-weighted variant $\sum_s\max_g n(s,g)\,/\,T$. High purity means clone states correspond cleanly to single physical places.

\textbf{Projected map and edge F1.}
Latent transitions are projected onto a place graph by
\[
W(g,g')=\max_{a}\;\max_{\substack{i,j\in\mathcal{V}\\ \chi(i)=g,\;\chi(j)=g'}}T_{a,i,j},
\]
and thresholded into a learned edge set $\widehat{\mathcal{E}}(\eta)=\{(g,g'):g\neq g',\,W(g,g')>\eta\}$. With $\mathrm{tp}=\lvert\widehat{\mathcal{E}}\cap\mathcal{E}\rvert$, precision, recall and F1 are defined in the usual way. We report F1 over a threshold sweep $\eta\in\{0.01,0.05,0.1,0.2,0.3\}$.

\textbf{Viterbi-path map and edge F1.}
The projected map above thresholds \textit{every} learned transition and therefore admits weak false edges. As a sharper read-out we keep only the transitions the decoded path actually takes: from $s^\star_{1:T}$ we tally consecutive state transitions, project each to a place edge $(\chi(s^\star_t),\chi(s^\star_{t+1}))$ with $\chi(s^\star_t)\neq\chi(s^\star_{t+1})$, and retain an edge once it is traversed in more than a small fraction of the episode (we use $0.2\%$ of the $T$ steps). Precision, recall and F1 against $\mathcal{E}$ are then computed as above. This Viterbi-path read-out is the ``Map F1'' reported in Table 3 and the Viterbi-path columns of Table 4 (Section 5.3).

\textbf{Action-next-cell accuracy.}
For every represented place $g$ and action $a$, the predicted next place $\arg\max_{g'}W_a(g,g')$ is compared with the environment's true successor $\mathrm{next}(g,a)$.

\textbf{Token--place entropies.}
From the empirical token/place co-occurrence we report $H(\text{token}\mid\text{place})$ and $H(\text{place}\mid\text{token})$; the former is small when perception is consistent, the latter reflects the (irreducible) aliasing of the environment.

\subsection{Implementation and hyperparameters}
The encoder is composed of three convolutional layers with strides $2,2,1$, followed by global average pooling and a dense projection to $\mathbb{R}^{D}$; the decoder uses the corresponding mirrored architecture. The gradCSCG forward pass, soft forward pass, and training steps are implemented as compiled TensorFlow graphs. Viterbi decoding is performed outside the compiled graph during inference. Transition logits are initialized with a bias toward the sink state so probability mass is well-defined before training. Table 1 lists all hyperparameters.

\begin{table}[t]
\caption{Hyperparameters. Ranges span the four environments of Section 4.}
\centering\small
\begin{tabular}{@{}l c c@{}}
\toprule
Component / parameter & Symbol & Value \\
\midrule
\multicolumn{3}{@{}l}{\textit{\textbf{VQ-VAE}}}\\
Codebook size & $K$ & 4--10 \\
Embedding dimension & $D$ & 32 \\
Encoder base width & --- & 32 \\
Commitment weight & $\beta$ & 0.25 \\
EMA decay & $\gamma$ & 0.99 \\
EMA smoothing & $\epsilon$ & $10^{-5}$ \\
\multicolumn{3}{@{}l}{\textit{\textbf{gradCSCG (cloned HMM)}}}\\
Action alphabet & $A$ & 4 \\
Clones per token & $C_k$ & 4--20 \\
Latent states & $N$ & 21--151 \\
\multicolumn{3}{@{}l}{\textit{\textbf{Joint training}}}\\
Softmax temperature & $\tau$ & 1.0 \\
gradCSCG-loss weight & $\lambda$ & 1.0 \\
Anneal horizon & $T_{\mathrm{anneal}}$ & iters/4 \\
Diversity weight & $\alpha_{\mathrm{div}}$ & 0.1 \\
gradCSCG learning-rate factor & --- & 100 \\
Chunk length & $T$ & 256 \\
Joint minibatch & $\lvert\mathcal{B}\rvert$ & 4 \\
Joint iterations & --- & 2000--5000 \\
\multicolumn{3}{@{}l}{\textit{\textbf{Finalization}}}\\
Iterations & --- & 1000--5000 \\
Transition-entropy weight & $\eta$ & $10^{-3}$ \\
Minibatch & --- & 8 \\
\multicolumn{3}{@{}l}{\textit{\textbf{Optimization and data}}}\\
Optimizer & --- & Adam [19] \\
VQ-VAE / joint learning rate & --- & $3\times10^{-4}$ \\
Finalization learning rate & --- & $10^{-2}$ \\
Episodes $\times$ steps & --- & $(4\text{--}10)\times10{,}000$ \\
\bottomrule
\end{tabular}
\end{table}

\subsection{Environments}

Our experiments take place in a family of controlled navigation environments. Each environment is a set of discrete locations --- \textit{places} --- connected into a graph that is the hidden ground-truth map; in this work the graph is a 2-D grid, so each place is a cell linked to its immediate neighbours. An agent explores by a \textbf{random walk}: at every step it occupies one place, receives an \textit{observation} produced by that place, and takes one of four movement actions (up, down, left, right) that moves it to an adjacent place, with walls and boundaries sticky (an invalid move leaves it where it is). Everything the learner ever sees is this stream of alternating observations and actions; the agent's true location and the graph's adjacency are never exposed, and are kept only for evaluation. The task is to reconstruct the map --- which places border which --- from the stream alone.

This is exactly the setting in which the original CSCG was validated, with one simplification on the perception side: there, each place emits a \textit{single discrete symbol} from a small alphabet, so the observation vocabulary is given in advance [9]. We keep everything else --- the known ground-truth topology, the strong perceptual aliasing, and the hidden position --- but replace the symbol at each place with a \textit{raw image}, so the vocabulary is no longer given and must be learned from pixels. This lets us ask, on the very maps George et al.\ used, whether a cognitive map can still be recovered once perception is itself part of the learning problem.

Formally, write $v_t\in\mathcal{V}$ for the place visited at time $t$, $a_t\in\{\text{up},\text{down},\text{left},\text{right}\}$ for the action taken, and $x_t$ for the observation received; an episode is the resulting stream
\[
(x_1,a_1),(x_2,a_2),\dots,(x_T,a_T).
\]
Each cell $v$ carries a digit label $d(v)\in\{0,\dots,9\}$, and we use two observation models:
\begin{itemize}[leftmargin=1.4em,itemsep=2pt,topsep=2pt]
\item \textbf{Symbolic} (the original-CSCG control [9]): the observation is the digit token itself, $x_t = d(v_t)$.
\item \textbf{Image} (our benchmark): the observation is a freshly drawn MNIST image of that digit, $x_t \sim \mathcal{D}_{d(v_t)}$, where $\mathcal{D}_{d}$ is the empirical MNIST distribution for digit $d$.
\end{itemize}
Under the image model, repeated visits to one cell never produce identical pixels (non-stationary appearance), while distinct cells that share a digit produce visually similar observations (aliasing). The model must therefore turn high-dimensional, variable images into \textit{stable} discrete tokens before action context can resolve the remaining aliasing --- a strictly harder problem than the symbolic one, and a more realistic test of whether temporal-structure learning can be coupled to learned perception.

The four environments (Table 2; Figure 2) are MNIST analogues of the canonical demonstrations of the original CSCG [9], each isolating a different facet of map-learning under aliasing:
\begin{itemize}[leftmargin=1.4em,itemsep=2pt,topsep=2pt]
\item \textbf{\texttt{aliased}} --- a $4\times4$ room with four digit classes arranged so that each recurs four times. With only four distinct observations across sixteen places, appearance alone is almost uninformative; this mirrors George et al.'s room with four unique observations (Fig.\ 2a,b of [9]).
\item \textbf{\texttt{corridors}} --- a $5\times5$ layout whose interior walls carve narrow corridors joined by repeated digits, a walled-maze variant that stresses recovery through bottlenecks.
\item \textbf{\texttt{room}} --- a $6\times6$ room in which a ring of distinct border digits surrounds a $4\times4$ interior of a \textit{single} repeated digit, so sixteen interior cells look identical. This mirrors George et al.'s uniform-interior room (Fig.\ 2c,d of [9]); the large aliased core is the hardest case for a fixed clone budget, and we use it to illustrate that the per-observation clone budget can simply be set by hand (Section 5.3).
\item \textbf{\texttt{two\_rooms}} --- a $13\times9$ map of two offset rooms that share a $3\times3$ patch, so one local appearance occurs at two globally distinct places (a \textit{confounder}). This mirrors the two overlapping rooms George et al.\ use to probe transitive inference (Fig.\ 2e,f of [9]); it is the largest and most aliased benchmark.
\end{itemize}

\begin{table*}[t]
\caption{Benchmark environments. $K$ is the number of discrete observation tokens (codebook size); ``Places'' counts walkable cells; the last column names the corresponding experiment in the original CSCG paper [9].}
\centering\small
\begin{tabular}{@{}l c c c l@{}}
\toprule
Environment & Grid & Places & Tokens $K$ & Original-CSCG analogue [9] \\
\midrule
\texttt{aliased}   & $4\times4$  & 16 & 4  & room, four unique observations (Fig.\ 2a,b) \\
\texttt{corridors} & $5\times5$  & 19 & 6  & walled maze \\
\texttt{room}      & $6\times6$  & 36 & 10 & uniform-interior room (Fig.\ 2c,d) \\
\texttt{two\_rooms}& $13\times9$ & 87 & 10 & two overlapping rooms (Fig.\ 2e,f) \\
\bottomrule
\end{tabular}
\end{table*}

\begin{figure*}[t]
\centering
\includegraphics[width=0.98\textwidth]{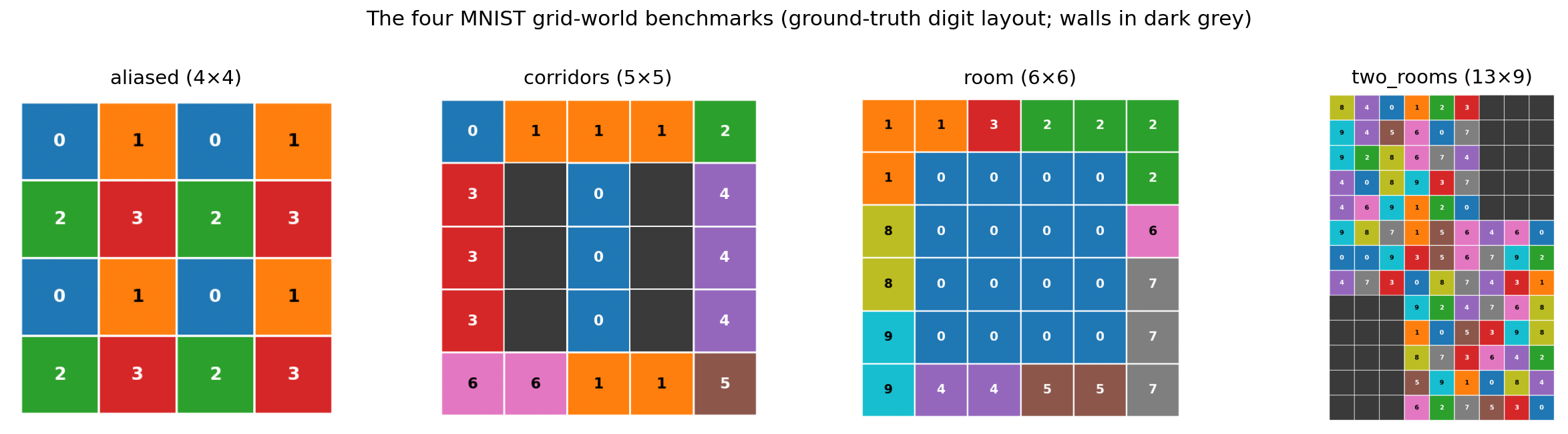}
\caption{The four benchmark environments, each shown as its grid of per-cell digit classes (walls in grey). Cells that share a digit are perceptually aliased; in the image benchmark every visit to a cell returns a different MNIST sample of its digit. Note the uniform interior of \texttt{room} and the shared corner that links the two halves of \texttt{two\_rooms}.}
\end{figure*}

For each environment we collect action-conditioned uniform random walks, 4-10 episodes of 10,000 steps, and never expose the agent's position or the adjacency graph to the model; both are retained only to score a held-out episode with the metrics of Section 3.9. Each run executes the three-stage pipeline of Section 3.6 (VQ-VAE warm-up $\rightarrow$ joint training $\rightarrow$ finalization). During warm-up an optional, benchmark-only digit-classification loss may be applied to the encoder; since object classes are unknown in a general environment it is disableable, and the perceptual front-end learns to discretize the observations without it.

\section{Results}
We evaluate gradCSCG on the four grid-worlds of Section 4. Training consumes only the observation--action stream; the agent's position and the true adjacency graph are withheld and used solely to score a held-out episode (Section 3.9). The argument runs in three steps: gradient training reproduces the cognitive maps of the original, EM-trained CSCG (Section 5.1); that recovery survives the replacement of the given symbol by a raw, never-repeating image (Section 5.2); and the gradient-trained formulation affords further properties of practical value (Section 5.3). Table 3 summarises the headline metrics; the subsections establish each claim in turn.

\begin{table*}[t]
\caption{Main results across the four environments (image observations; \texttt{room} uses a hand-set per-observation clone budget). State-to-place purity is visit-weighted; map recall and F1 are read from the Viterbi-path graph (Section 5.3, Table 4), with recall 1.00 throughout. Entries are means over repeated runs; the seed-to-seed s.d.\ is $\le0.01$ for every environment except \texttt{two\_rooms} ($\le0.03$).}
\centering\small
\setlength{\tabcolsep}{5pt}
\begin{tabular}{@{}l c c c c c c c c c@{}}
\toprule
Environment & Grid & $K$ & States $N$ & Perplexity & Clone purity & State$\rightarrow$place purity & Action acc.\ & Map recall & Map F1 \\
\midrule
\texttt{aliased}            & $4\times4$  & 4  & 21  & 4.00 & 0.98 & 0.98 & 1.00 & 1.00 & \textbf{1.00} \\
\texttt{corridors}          & $5\times5$  & 6  & 31  & 5.71 & 0.86 & 0.98 & \textbf{1.00} & 1.00 & 0.98 \\
\texttt{room} (hand-set clones) & $6\times6$ & 10 & 57 & 6.56 & 0.83 & 0.95 & 0.97 & 1.00 & 0.98 \\
\texttt{two\_rooms}         & $13\times9$ & 10 & 151 & 9.46 & 0.88 & 0.98 & 0.99 & 1.00 & \textbf{1.00} \\
\bottomrule
\end{tabular}
\end{table*}

\subsection{Gradient descent recovers the cognitive maps of the original CSCG}
\textbf{The learned latent graph contains the true map.} Trained only on observation--action walks, gradCSCG concentrates its transition probability on the edges that mirror the environment's physical adjacency, and map-edge \textbf{recall is 1.00} in every environment. This is the central result of the original CSCG [9], obtained here by back-propagation rather than expectation--maximization. We first reproduce it in the \textit{original symbolic setting} of [9], in which each cell emits its integer observation directly with no images: trained by gradient descent, gradCSCG recovers the same maps George et al.\ obtain by EM --- a square room laid out as a 2-D grid, a rectangular room whose uniform interior is distinguished only by its walls, and three disjoint rooms stitched into a single coherent graph (Figure 3, cf.\ Fig.\ 2 of [9]). The recovered transition graph reproduces the room topology, and the node colours (one per observation) show that perceptually identical cells are split into separate, correctly-placed clones. This isolates the change of training rule from the change of observation model; the same recovery then carries over to raw images (Section 5.2).

\begin{figure*}[t]
\centering
\includegraphics[width=0.98\textwidth]{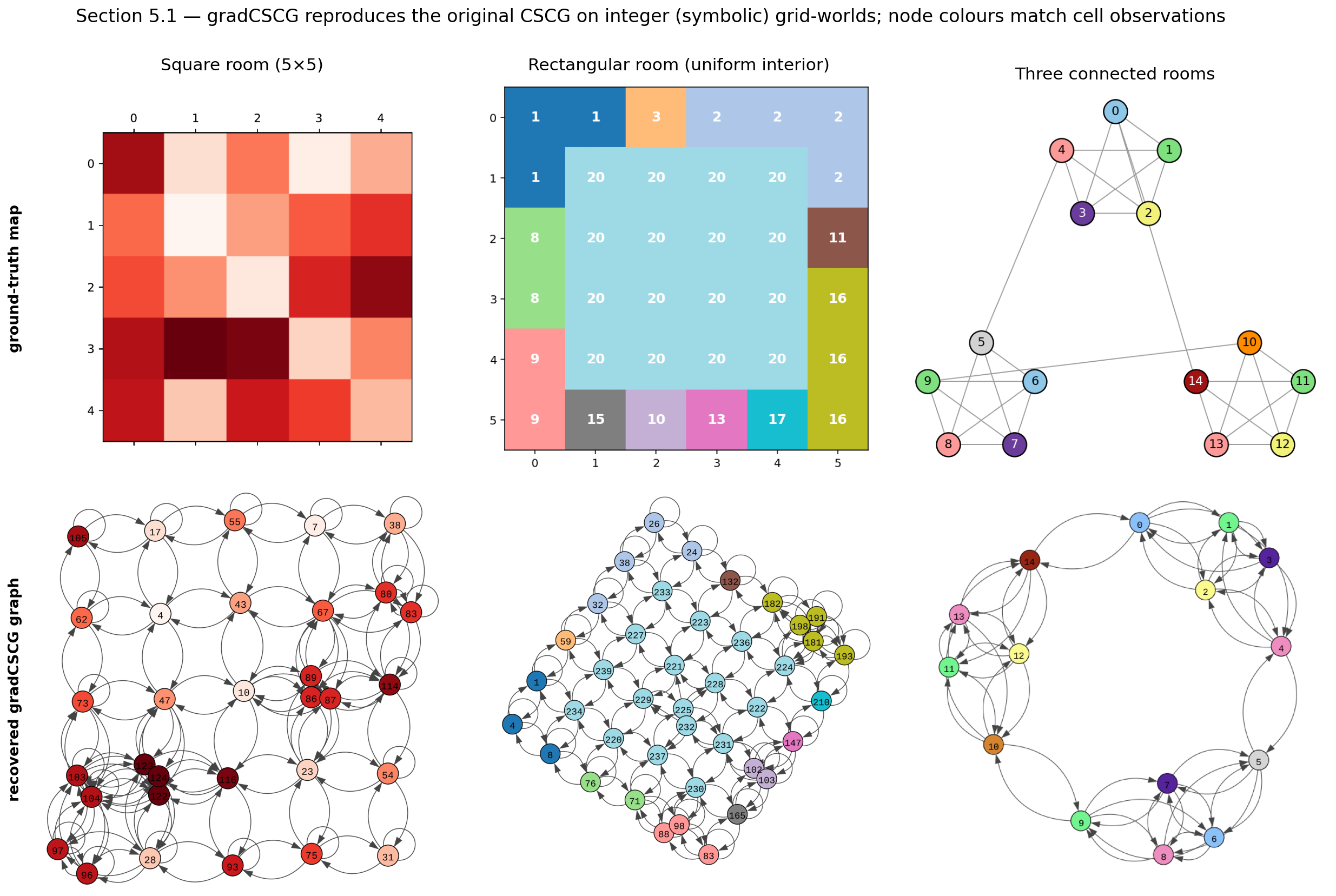}
\caption{gradCSCG reproduces the original CSCG on integer (symbolic) grid-worlds, trained by gradient descent rather than EM. \textbf{Top row:} the ground-truth map of each environment, every cell (or place) coloured by its observation. \textbf{Bottom row:} the transition graph recovered by gradCSCG, with nodes coloured by the observation they emit and positioned by the learned connectivity --- colours are matched cell$\leftrightarrow$node, so the correspondence is direct. \textbf{Left:} a square room recovered as a 2-D grid from a heavily aliased symbol stream (cf.\ George et al.\ [9], Fig.\ 2a,b). \textbf{Middle:} a rectangular room whose uniform interior --- the large block of a single repeated observation --- is disambiguated only by the surrounding walls (cf.\ Fig.\ 2c,d). \textbf{Right:} three pentagonal rooms with identical local observations, stitched into one coherent graph by transitive inference (cf.\ Fig.\ 2e,f). In every case the recovered graph matches the ground-truth topology.}
\end{figure*}

\textbf{Clones absorb the aliasing into place-specific states.} Recovering topology from aliased observations requires splitting one observation into context-specific latent states --- the clone mechanism. Decoding a held-out episode and assigning each visited clone to its majority place, clones are almost perfectly place-specific: visit-weighted state-to-place purity is 0.95--1.00, every place is represented by at least one clone, and a typical place is covered by only 1.5--1.7 clones (Table 3; the clone-in-layout panels of Figure 5). Gradient training therefore reproduces the context-split, place-cell-like representations that make the original CSCG interpretable.

\textbf{The recovered map answers ``where does this action lead?''.} A map is useful only if it can be queried for the outcome of an action. Reading the most likely successor place for each (place, action) pair off the learned transitions and comparing it with the true successor, action-outcome accuracy is 0.97--1.00 (Table 3) --- the graph is consistent enough to support the one-step, inference-based planning of [9].

\subsection{The maps survive raw, non-stationary, aliased images}
\textbf{A learned front-end supplies CSCG-grade tokens from pixels.} Where the original CSCG is handed clean symbols, our model must manufacture them. The VQ-VAE does so almost deterministically: a place emits a single token on 1.0--1.1 of its visits on average, the conditional entropy $H(\text{token}\mid\text{place})$ is only 0.09--0.18 nats, and codebook perplexity tracks the number of digit classes --- even though that place is never seen twice. Reconstructions stay faithful throughout joint training (Figure 4), confirming that the topological objective sharpens rather than collapses the codebook.

\textbf{Topology recovery holds end-to-end and at scale.} Driving gradCSCG with these \textit{learned} tokens --- not the ground-truth digits --- the full pixel-to-map pipeline still recovers every adjacency (recall 1.00) up to the largest, most aliased environment, \texttt{two\_rooms} ($13\times9$, 87 places, 151 latent states), at state-to-place purity 0.98 and map F1 1.00 (Figure 5; Table 3). The map that George et al.\ recover from symbols, we recover from images.

\begin{figure*}[t]
\centering
\includegraphics[width=0.98\textwidth]{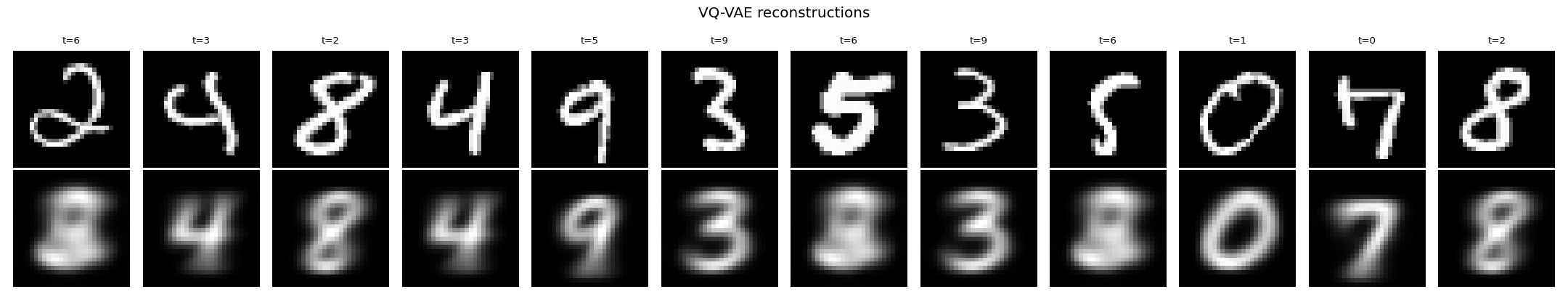}
\caption{Perception is stable under joint training (\texttt{two\_rooms}). Top: input MNIST observations; bottom: VQ-VAE reconstructions, each annotated with the discrete token assigned to it. Every visit to a place yields a different image, yet the assigned token is consistent.}
\end{figure*}

\begin{figure*}[t]
\centering
\includegraphics[width=0.99\textwidth]{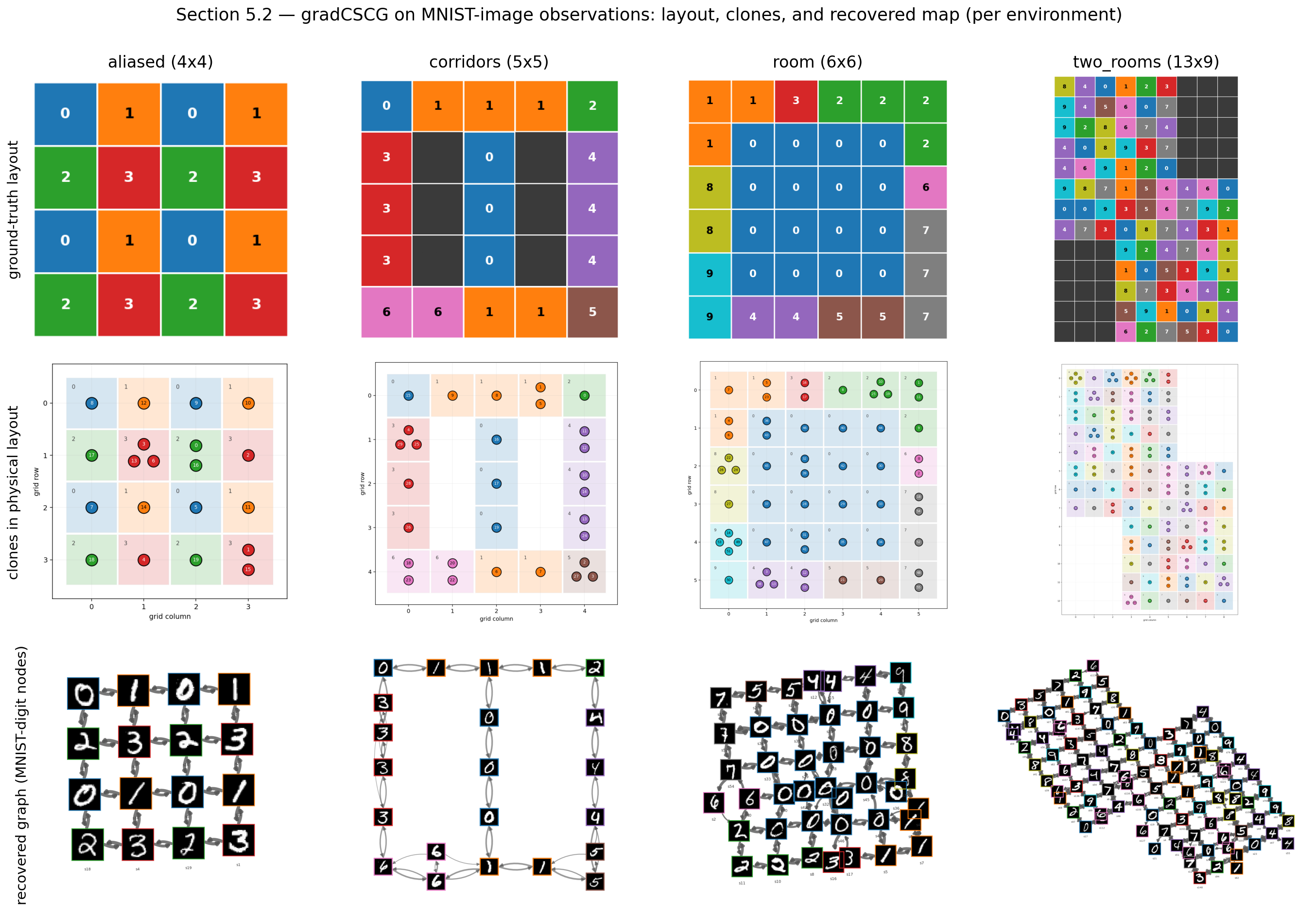}
\caption{gradCSCG on raw MNIST observations, one column per environment (\texttt{aliased}, \texttt{corridors}, \texttt{room}, \texttt{two\_rooms}). \textbf{Top row:} ground-truth layout (cells coloured by digit class). \textbf{Middle row:} the decoded clone states drawn at their assigned physical cells (transitions omitted for clarity) --- perceptually identical cells (same colour, repeated across the grid) are split into separate clones that sit at the correct places. \textbf{Bottom row:} the recovered transition graph, nodes drawn as the MNIST digit they emit. The topology is recovered from images in every environment, up to the $13\times9$ \texttt{two\_rooms}.}
\end{figure*}

\subsection{Additional properties of the gradient-trained model}
\textbf{A minor convenience: per-observation clone budgets.} The clone count is a hyperparameter that can be set per observation rather than uniformly. Giving the heavily-aliased interior digit of \texttt{room} more clones (20) than its rarely-repeated border digits (4) recovers the same map with 57 latent states instead of 201, at no cost to recall. This is a \textit{static, hand-set} allocation fixed before training --- not a learned, on-demand assignment of clones --- so we note it only as a practical knob, not a contribution.

\textbf{Reading the map off the Viterbi path recovers it almost exactly.} Projecting \textit{every} above-threshold latent transition into the place graph floods it with weak false edges, so the naive precision is low (0.16--0.44 at $\eta=0.01$; Table 4, left). But those weak edges are decoding noise, not routes the agent ever takes. We therefore keep only the place edges that the Viterbi-decoded path actually traverses more than a small fraction of the episode --- \textbf{0.2\% of the $T$ steps} (20 transitions at $T=10{,}000$). This is a \textit{minimum-traffic} cutoff that asks for genuine, repeated movement rather than a single mis-step, and it scales with episode length instead of being a fixed magic number. It removes essentially all false edges while leaving recall untouched: precision rises to \textbf{0.95--1.00} and map F1 to \textbf{0.98--1.00} across all four environments, at recall 1.00 (Table 4). The read-out is robust to the exact fraction: because true adjacencies are traversed hundreds of times per episode and spurious ones only a handful, any cutoff from $\sim$0.1\% to $\sim$0.5\% of $T$ (10--50 transitions here) leaves recall at 1.00 while precision saturates by $\sim$0.2\%.

\begin{table*}[t]
\caption{Map-edge scores from the projected transition graph (every edge with weight $>0.01$) versus the Viterbi-path graph (edges traversed in more than 0.2\% of the $T$ steps, i.e.\ $>20$ at $T=10{,}000$). Recall is 1.00 in both columns; the Viterbi-path read-out removes the weak false edges that depress the projected-graph precision.}
\centering\small
\begin{tabular}{@{}l c c c c c@{}}
\toprule
Environment & Projected $P$ & Projected $F_1$ & Viterbi-path $P$ & Viterbi-path $R$ & Viterbi-path $F_1$ \\
\midrule
\texttt{aliased}            & 0.44 & 0.62 & \textbf{1.00} & 1.00 & \textbf{1.00} \\
\texttt{corridors}          & 0.24 & 0.39 & \textbf{0.95} & 1.00 & \textbf{0.98} \\
\texttt{room} (hand-set clones) & 0.27 & 0.42 & \textbf{0.95} & 1.00 & \textbf{0.98} \\
\texttt{two\_rooms}         & 0.16 & 0.27 & \textbf{1.00} & 1.00 & \textbf{1.00} \\
\bottomrule
\end{tabular}
\end{table*}

\textbf{Joint training is stable.} Through the joint phase the reconstruction loss stays flat while the length-normalized gradCSCG term falls once its weight $\lambda_t$ ramps in; the finalization phase then sharply reduces the gradCSCG NLL with the encoder frozen (Figure 6) --- the behaviour the loss-balancing terms of Section 3.6 are designed to produce.

\begin{figure*}[t]
\centering
\includegraphics[width=0.98\textwidth]{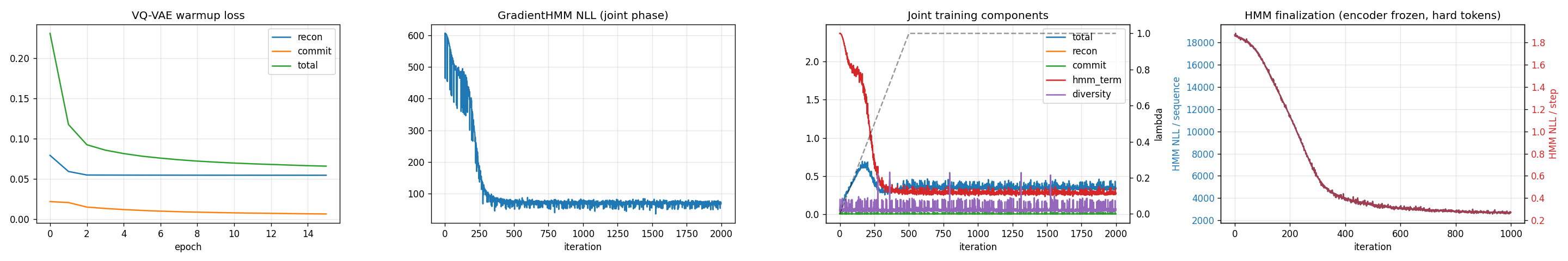}
\caption{Training dynamics (\texttt{two\_rooms}). VQ-VAE warm-up, the joint phase (reconstruction, commitment, length-normalized gradCSCG, and diversity terms), and the pure-gradCSCG finalization phase with the encoder frozen.}
\end{figure*}

\section{Discussion}
Our central result is a \textbf{proof of principle}: the Clone-Structured Cognitive Graph (CSCG), a normative account of how the hippocampus builds cognitive maps [9, 10], can be reformulated to live inside a gradient-trained pipeline. Re-deriving its forward algorithm as a differentiable, log-space computation lets the model be optimized by backpropagation rather than expectation--maximization.
On the original symbolic grid-worlds the gradient-trained model reproduces the hallmark behaviour of George et al.\ --- recovering room topology from heavily aliased observations and splitting identical-looking places into context-specific clones (Section 5.1).
Crucially, the reformulation also makes CSCG \textbf{composable}: co-trained with a VQ-VAE perceptual front-end, it recovers cognitive maps of pixel-based environments --- where every place looks different on each visit and many places look alike --- that the symbolic model cannot even ingest (Sections 5.2--5.3).
The contribution is therefore less a new map-learning result than a change of \textit{substrate}: CSCG becomes a module that can be wired into, and trained jointly with, neural networks.

This change of substrate is worth making because the two research traditions that bear on map learning each pay a price the other avoids.
Interpretable cognitive-map models --- CSCG [9, 10], the Tolman--Eichenbaum Machine [14], the successor/predictive map [3] --- recover sparse, relational structure that supports planning and generalization, but they presuppose a discrete, \textit{given} observation alphabet and so cannot, by themselves, look at pixels.
Neural world models --- the Dreamer family of recurrent state-space models [6] and the vector-quantized representation learners they build on [12] --- consume raw perception readily, but their latent dynamics are dense and hard to read as a map.
gradCSCG sits in the gap between them: it keeps CSCG's sparse, aliasing-resolving clone structure while delegating perception to a learned front-end.
The enabling fact is not itself new --- the HMM forward recursion is a differentiable computation [16], and neural emissions for HMMs have been studied in language modelling [17] --- but applying it to the \textit{action-augmented} cloned HMM, and letting the topological objective propagate back to shape perception, is what turns an ordinary encoder into a \textit{sequence-aware} one.

That coupling is not free, and the way it fails is instructive.
Trained naively, the joint objective collapses the codebook, because the sequence likelihood grows with episode length and swamps reconstruction; the loss-balancing of Section 3.6 --- length normalization, weight annealing, and a diversity penalty --- is what makes the two modules cooperate rather than compete, and in our experiments it is the difference between learning a map and learning a single token.
We regard this as the main practical lesson for any future CSCG-plus-network hybrid: the perceptual and sequence objectives must be explicitly balanced, or the stronger one consumes the other.

\section{Limitations and Outlook}
\textbf{Limitations.} Two caveats bound the present results directly.
\textit{(i) Read-out threshold.} High precision requires reading the map off the Viterbi-decoded path with a minimum-traffic cutoff (Section 5.3); the cutoff is adaptive (a fixed fraction of episode length) and recall is insensitive to it, but a fully parameter-free read-out --- e.g.\ stronger transition-entropy regularization or edge calibration --- is still open.
\textit{(ii) Scope.} The environments are controlled, static, 2-D MNIST grid-worlds; we have not yet tested 3-D, partially observed, non-stationary, or visually richer settings.

\textbf{Outlook.}
The deeper limitation is conceptual, and it points to where this work goes next.
By construction, CSCG captures the \textit{static, relational} skeleton of an environment --- which places exist and how they connect --- but \textbf{true behavioural flexibility needs more than a static map}.
A realistic agent must also track the \textit{dynamic} elements of a scene (objects that move, appear, or change), handle far more complex and higher-dimensional observations, and adapt \textit{quickly} as the world changes --- none of which a clone graph over fixed places is designed to do.
We therefore see gradCSCG not as a complete agent but as the \textit{structural} component of a larger system, with genuine flexibility coming from pairing it with complementary modules trained in the same differentiable framework.
The most natural partner is a \textbf{recurrent network (RNN)}: where the clone graph holds the slow, stable map, an RNN can flexibly carry the fast-changing, dynamic context the map omits [6] --- a division of labour that mirrors how real neural circuits combine stable spatial codes with rapidly updating population activity.
Richer input --- 3-D or egocentric vision --- would in turn call for a heavier perceptual front-end that pre-processes raw views before they are discretized.
Making CSCG gradient-trained is precisely what makes such hybrids buildable, since every part can then be optimized by the same backpropagation.

Many questions remain open.
The most immediate is \textbf{dynamic clone recruitment}: here the clone budget is a static hyperparameter fixed before training (Section 5.3), whereas the cloning principle ultimately calls for \textit{recruiting a new clone whenever a novel context appears}.
A differentiable mechanism that grows or prunes clones on demand would let the model size itself to an environment's true complexity --- and learn it faster and more incrementally --- a natural next step now that training is gradient-based.
Threshold-free map read-out, scaling to larger and partially observed worlds, and coupling the learned map to a planning or reinforcement-learning loop [6] are further directions.

\section{Conclusion}
We have shown that the Clone-Structured Cognitive Graph can be trained by gradient descent while faithfully reproducing the behaviour of its EM-trained original, and that, so trained, it can be grounded in a learned VQ-VAE front-end to recover cognitive maps of pixel-based environments.
We close on what this is \textit{for}.
The hippocampus --- of which CSCG is a normative model [9, 10] --- does not work in isolation; it operates in close interaction with the entorhinal cortex, prefrontal cortex, and many other regions [14], each carrying out a different kind of computation, and the cognitive map is useful precisely because it is embedded in that larger system.
We read our contribution in the same spirit: a gradient-trained CSCG is not a monolithic solution but a \textit{composable} module, and rephrasing it in the language of deep learning is what lets it be wired together with the complementary machinery it needs --- neural perception, recurrent tracking of dynamics, and planning.

Because our formulation is differentiable, the benefit runs in both directions: the same module that provides explicit representations to artificial agents, may also become a more scalable normative model for studying how the brain builds and uses such representations.
Unlocking the power of world models, in brains and in machines alike, is therefore less about the module on its own than about a clever division of labour among specialized parts.

\textbf{Code and data availability.} The implementation, benchmark environments, training scripts and evaluation suite are available in the project repository.

\section*{Code and Data Availability}
The complete implementation of the \texttt{gradCSCG} pipeline, VQ-VAE discretization modules, synthetic training environments, and evaluation suite is publicly available on GitHub at \url{https://github.com/tristanstoeber/gradCSCG}. The MNIST datasets used in this work are open-source and automatically fetched by the provided training scripts.

\section*{Acknowledgements}
AN, PG acknowledge support by the Goethe Research Exchange Program. TMS acknowledges funding from the “Advancing World model Learning with Neural Cloned-Structured Causal Graphs” project, funded by the European Union, via the oc1-2024-TES-01 issued and implemented by the ENFIELD project, under the grant agreement No 101120657; and from the CIDAS Fellowship 2025, University of Göttingen.

\end{document}